\documentclass[conference]{IEEEtran}
\usepackage{bm}
\usepackage{graphicx}
\usepackage{amsmath}
\usepackage{indentfirst}
\usepackage{bm}
\usepackage{enumerate}

\usepackage{multirow} 
\usepackage{cite}
\usepackage{array}
\usepackage{amsmath,amssymb,amsfonts}
\usepackage{algorithm,algorithmic}
\usepackage{graphics}
\usepackage{epsfig}
\usepackage[tight,footnotesize]{subfigure}
\usepackage{enumerate}
\usepackage{pifont}
\usepackage[english]{babel}
\usepackage{xparse}				
\usepackage{multirow, booktabs} 
\usepackage[table]{xcolor}		
\usepackage{verbatim}

\ifCLASSINFOpdf
\else
\fi
\begin{document}
%
\title{Neural Population Coding for \\ Effective Temporal Classification}


\IEEEoverridecommandlockouts 
\author{\IEEEauthorblockN{Zihan Pan$^{\ast}$, Jibin Wu$^{\ast}$\thanks{* Zihan Pan and Jibin Wu contributed equally to this work.}, Malu Zhang, Haizhou Li}
\IEEEauthorblockA{Department of Electrical and Computer Engineering\\
National University of Singapore\\
Singapore, 117583\\
}
\and
\IEEEauthorblockN{Yansong Chua$^{\dagger}$\thanks{$\dagger$ Corresponding Email: chuays@i2r.a-star.edu.sg}}
\IEEEauthorblockA{Institute for Infocomm Research\\
Agency for Science, Technology and Research\\
Singapore, 138632\\
}
}


%


\maketitle

\begin{abstract}
Neural encoding plays an important role in faithfully describing the temporally rich patterns, whose instances include human speech and environmental sounds. For tasks that involve classifying such spatio-temporal patterns with the Spiking Neural Networks (SNNs), how these patterns are encoded directly influence the difficulty of the task. In this paper, we compare several existing temporal and population coding schemes and evaluate them on both speech (TIDIGITS) and sound (RWCP) datasets. We show that, with population neural codings, the encoded patterns are linearly separable using the Support Vector Machine (SVM). We note that the population neural codings effectively project the temporal information onto the spatial domain, thus improving linear separability in the spatial dimension, achieving an accuracy of 95\% and 100\% for TIDIGITS and RWCP datasets classified using the SVM, respectively. This observation suggests that an effective neural coding scheme greatly simplifies the classification problem such that a simple linear classifier would suffice. The above datasets are then classified using the Tempotron, an SNN-based classifier. SNN classification results agree with the SVM findings that population neural codings help to improve classification accuracy. Hence, other than the learning algorithm, effective neural encoding is just as important as an SNN designed to recognize spatio-temporal patterns. It is an often neglected but powerful abstraction that deserves further study.

\end{abstract}


%
\IEEEpeerreviewmaketitle



%
\IEEEpeerreviewmaketitle

\section{Introduction}
The information is transmitted and exchanged using action potentials or spikes in the human brain. As such, biologically realistic spiking neural networks (SNN) have been proposed in the field of computational neuroscience to study the working principles of neural circuits and brain regions\cite{dayan2001theoretical}. Recently, the algorithmic power offered by deep artificial neural networks and low-power, highly parallel computation demonstrated by the human brain has attracted growing research attention on such biologically plausible computing models\cite{pfeiffer2018deep}\cite{zhang2018highly}\cite{zhang2018feedforward}. 

Fundamentally, the spiking neuron receives incoming spikes from its afferent neurons via dendrites; generates an action potential (or spike) and delivers to post-synaptic neurons when its membrane potential exceeds the firing threshold \cite{gerstner2002spiking}. The rich neuronal dynamic endows spiking neurons with the capability of using precise spike timing and phase for sparse coding and efficient computation. Several studies in neuroscience suggest that the low latency in our visual and auditory systems can be attributed to the precise timing of these action potentials \cite{butts2007temporal}\cite{crouzet2010fast}\cite{carr1990circuit}\cite{thorpe2001spike}. Hence, in this paper, we primarily focus our study on temporal neural codes based on precise spike timing, as opposed to rate coding that has a higher latency \cite{brette2015philosophy}.

The SNN models have been successfully applied for many pattern recognition tasks, instances include image classification \cite{tavanaei2018deep}, automatic speech recognition (ASR) \cite{pan2018event, wu2018biologically,tavanaei2017spiking,zhang2015digital} and environmental sound classification \cite{wu2018spiking,dennis2013temporal,xiao2018spike}. For auditory classification tasks, the time domain signals are usually pre-processed and converted into spatial-temporal spectrograms using Short-Time Fourier Transform (STFT). To further process these spectrograms with the SNN, a proper neural encoding scheme is required to convert real-valued input signals into spike patterns. For this purpose, many biologically inspired coding schemes have been proposed and studied, including latency, phase and threshold coding \cite{gerstner2002spiking}. 


Currently, most research works fail to consider the SNN learning system as a whole; while focusing on the learning algorithms used for the back-end SNN classifier, without giving sufficient attention to the encoding front-end. For example, in \cite{dennis2013temporal} \cite{xiao2016spiking} simple latency code is exploited; \cite{loiselle2005exploration} uses phase coding; \cite{zhang2017supervised}\cite{zhang2018empd} demonstrate the learning capacity of the algorithms by Poisson random patterns; population of latency codings is used in \cite{wu2018spiking}. They have achieved satisfying results, however, there are not sufficient reasons and discussion on the choices of neural encodings. Besides, we note, how inputs are encoded, could potentially simplify or complicate the learning task. Thus in this work, we look into the important research question: what is a good neural encoding scheme that allows an SNN to solve the speech recognition or sound classification tasks effectively? Here, given the several neural coding schemes available, we study which of these coding schemes may actually facilitate pattern recognition with an SNN classifier.

Motivated by the above, we make several contributions in this work: 1) we carry out a comparative study over the most commonly used neural encoding schemes: latency code, phase code, population-latency code, population-phase code and threshold code; 2) the biological plausibility of these encoding schemes are discussed; 3) we evaluate the efficacy of these coding schemes for the temporal pattern classification tasks and 4) lastly, we find that population code projects sparse spatial-temporal spike patterns onto extended spatial dimension such that they could be easily classified even with a linear classifier. Such spatially separated features could be easily learned by a simple SNN classifier. 


The rest of the paper is organized as follows: Section II. and Section III. present and review two classes of neural coding schemes: the temporal neural code of a single neuron and a population of neurons, respectively. Section IV. evaluates the efficacy of these coding schemes for both speech and sound recognition tasks. Section V. further discusses the characteristics of these neural coding schemes and we conclude the paper in Section VI.

\section{Neural temporal codes}
\label{Neural temporal codings}
The neural temporal codes record the dynamics of the external stimuli using the precise spike timing, as opposed to the average firing rate. Two commonly used such codes are the latency and phase code, which we will discuss in this section. These temporal codes may be applied to a single neuron or a population of neurons. Specifically, for the single neuron code, one neuron is used to encode a signal stream of stimulus (e.g., auditory stimulus from a single cochlear neuron). In contrast, for the population code, multiple neurons are utilized to encode a single stream.

\subsection{Latency Code}
Latency code, found for the first time in the human retina ganglion cells \cite{gollisch2008rapid}, has challenged the common understanding of human vision. It was previously believed that the average firing rate of neurons is the main underlying coding mechanism for visual stimuli. The latency code has also been found in animal auditory systems. One notable example is the bat auditory system, whereby echoes are encoded with precise timing for echolocation. It was shown that the latency code in auditory nerves represents both the monaural and binaural intensity cues induced by a head-related transfer function in the peripheral system \cite{fontaine2009bat}. The latency code has been exploited extensively for pattern recognition tasks, including both vision \cite{yu2013rapid} and auditory tasks \cite{pan2018event} \cite{wu2018biologically}. 

For latency code, the main assumption is that the encoding neuron will fire only once within each encoding time window. As we are only concerned about the latency of the first spike event, all subsequent spikes, if any, are ignored. Suppose we have a stream of stimuli, sampled and normalized as discrete time series $x=[x_1,x_2,...,x_N]$ at a sampling rate of $r$ samples per second. Here, $x_i$ refers to the signal intensity of time window $i$ and the value is normalized within $[0,1]$. The corresponding latency encoded spike train $t^f=[t_1, t_2,...,t_N]$ in seconds is defined as:

\begin{equation}
t_i= \frac{1}{r}(i-x_i)
\end{equation}

\begin{figure}[htbp]
    \centering
    \includegraphics[width=0.4\textwidth]{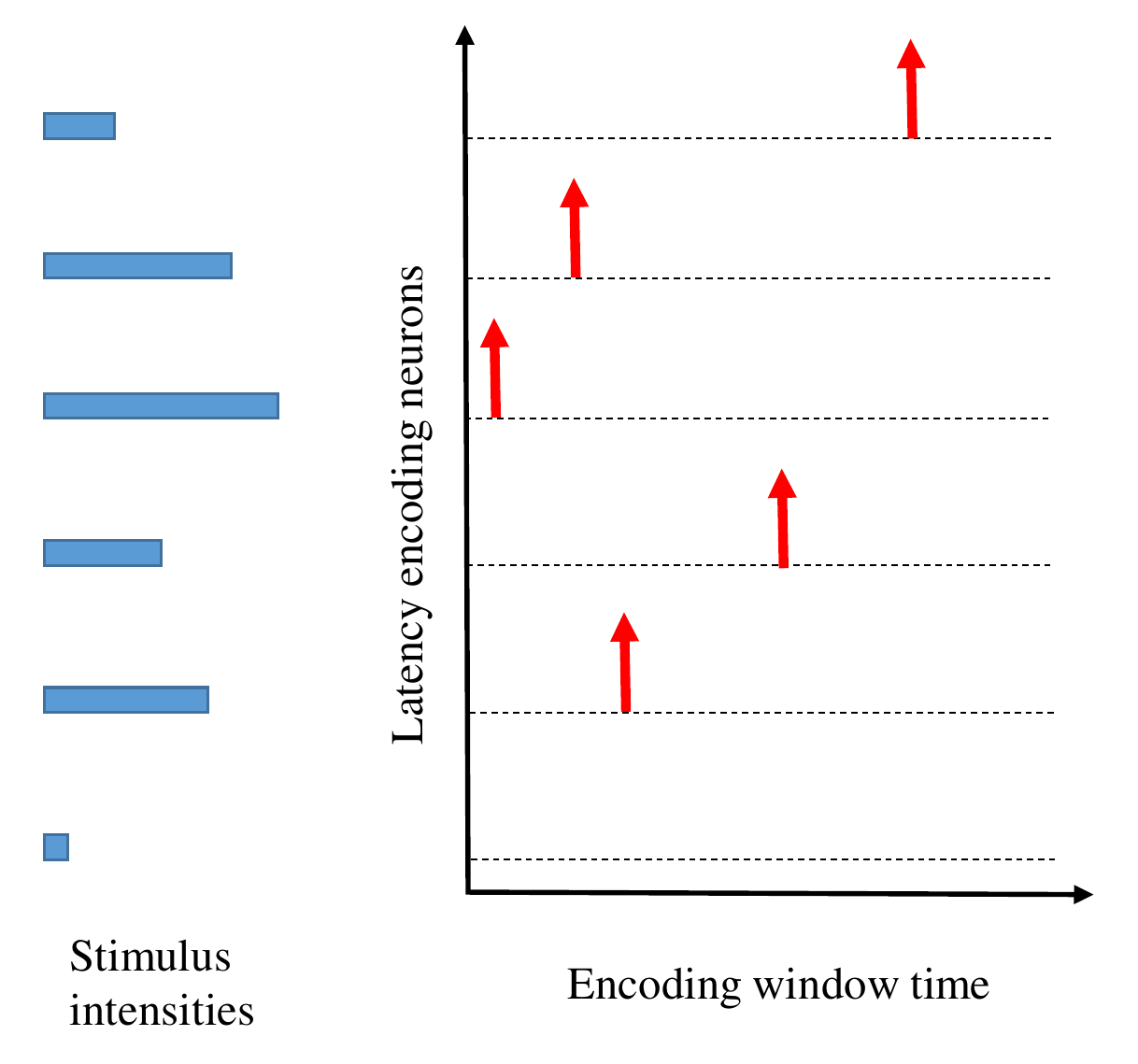}
    \caption{Latency coding: six different stimulus intensity levels (e.g., luminance in vision or sound pressure in audition, as shown using the horizontal blue bars on the left) are encoded in the latency of six different spikes of each encoding neuron, within a fixed encoding window. Each row represents one encoding neuron, and the red arrows indicate the spike timing. A more intense stimulus will generate an earlier spike. Furthermore, if the stimulus is too weak to trigger a spike, no spike will occur.}
    \label{latency code}
\end{figure}

As demonstrated in Figure \ref{latency code}, each spike timing $t_i$ encodes the intensity of the sensory stimulus that sampled within the particular time window $i$.

\subsection{Phase Code}
The transmission of sensory information within neural system should be robust to noise and the phase code has been introduced to improve the noise robustness \cite{nadasdy2009information}. The phase code relies on two basic assumptions: firstly, the spike times are phase-locked with the sub-threshold membrane potential oscillations (SMO) \cite{llinas1991vitro}; and secondly, the SMOs are synchronous with neighboring neurons \cite{benucci2007standing}. That is, SMOs share same frequencies while with a phase gradient among each other. Such phase gradients have been shown to exist in the visual cortex of the turtle \cite{prechtl2000direct}. Evidence for such phase-locking has also been observed in the human brains. For instance, during exploration of the surrounding environments \cite{chrobak1996high} and processing of sensory inputs \cite{robbe2006cannabinoids}, the high coherence among action potentials have been identified at the gamma (40-60 Hz) and theta (3-8 Hz) cycles. In auditory perception, cortical oscillations have also been observed \cite{giraud2012cortical} \cite{arnal2012cortical}.

\begin{figure}[htbp]
    \centering
    \includegraphics[width=0.4\textwidth]{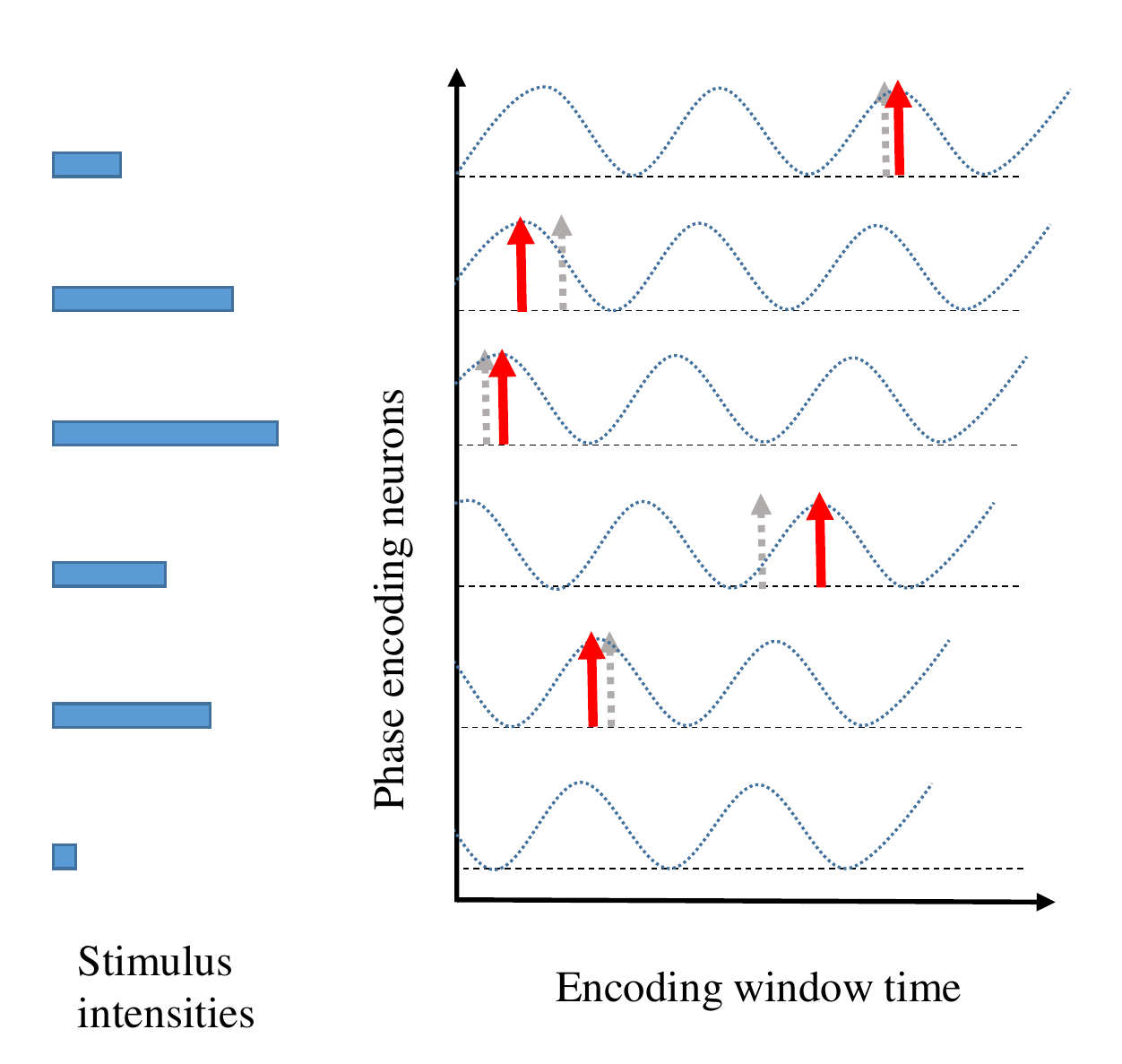}
    \caption{Phase coding: based on the latency code in Figure \ref{latency code}, spike timing on individual encoding neuron is synchronized with its sub-threshold membrane potential oscillation (SMO), which is shifted and phase-locked with the SMO of other neurons. Spike timing is set at the nearest peak of the SMO. Grey arrows are the latency coded spikes, as per Figure \ref{latency code}; red arrows are the actual phase-coded spikes.}
    \label{phase code}
\end{figure}

As illustrated in Figure \ref{phase code}, each encoding neuron has its own SMO (blue sinusoidal waves), which is phase-locked to the neighboring neurons with a constant phase shift. The preliminary spike timing is determined following latency code (grey dashed arrow) for each sampled stimuli, and then a spike is generated at the nearest peak of the sub-threshold membrane potential (red solid arrow). This process is defined as follows, 
\begin{equation}
    t^{p}_i = \delta (t_i, SMO_i)
\end{equation}
where the function $\delta (t_i, SMO_i)$ determines the nearest peak to the $t_i$ on the sub-threshold membrane potential. Here, $t_i$ denotes the preliminary spike timing as in Equation. (1) and $t^{p}_i$ denotes the phase-encoded spike timing of the $i_{th}$ sample. 

\section{Neural population codes}
While the above section on neural temporal codes describes coding schemes at the single neuron level, these coding schemes may, in general, be applied to a population of neurons. In this section, we describe how temporal code may be extended to population code, whereby multiple neurons are involved in encoding one signal stream of stimuli. In addition, we present the threshold code, a coding scheme unique to the population code.

\subsection{Population Latency/Phase Code}
It has been shown in both our visual and auditory systems, an external stimulus is perceived by a population of neurons; the receptive field of each neuron determines the part of the stimulus it is sensitive to \cite{eurich2000multidimensional} \cite{snippe1996parameter}. For example, in the visual cortex, each ganglion cell has its own receptive field, which defines the area of the visual field it is most responsive to \cite{gilbert1992receptive}. Similarly, in the auditory system, a population of neurons encode a vast range of sound levels with remarkable accuracies; in each frequency sub-band, a union of receptive fields from multiple neurons  will cover the full audible range \cite{dean2005neural}. 

\begin{figure}[htbp]
    \centering
    \includegraphics[width=0.45\textwidth]{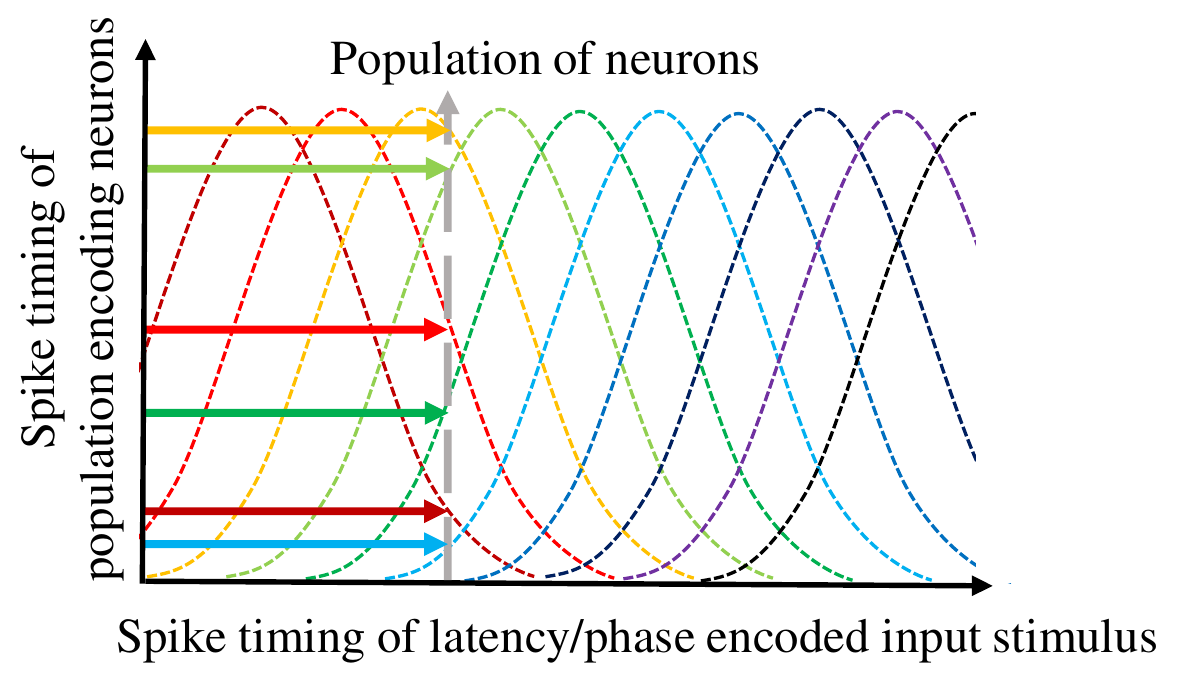}
    \caption{Population latency/phase coding: An input stimulus is being encoded by ten neurons. The x-axis refers to spike timing of the latency or phase coded input stimulus with each encoding neuron sensitive to only a certain range in this time window (as shown by the Gaussian receptive fields). The dashed grey arrow indicates the spike generated using the latency or phase code based on the stimulus value. The horizontal colored arrows indicate the neurons that encode the stimulus using the population code, with their respective spike timing determined by where the grey arrow (spike timing of input stimulus) intersects with their respective receptive fields.}
    \label{poplupation code}
\end{figure}

To implement the latency and phase code as described in Section 2 on a population of neurons, we extend the notion of the receptive field as described above to a population of neurons. In such a population, each neuron has an overlapped while distinct receptive field that as a whole covers the input range as shown in Figure \ref{poplupation code}. In this example, the latency encoded (Figure \ref{latency code}) or phase encoded (Figure \ref{phase code}) spike falls within the Gaussian receptive fields of six encoding neurons. These encoding neurons will emit a spike at the timing as determined by the y-axis value of the intersection points. The Gaussian receptive field was first introduced in \cite{bohte2002error} to encode the continuous input variables to spike time. It addresses the limited temporal precision and neuronal variability problems
of temporal code by distributing each input variable over multiple neurons.

\subsection{Threshold Code}
Information from the environment is almost always continuous, and time-varying. For example, spoken words in speech are continuous in time and may vary from tens to hundreds of milliseconds; even the duration of the same word may vary. Even visual perception of objects is continuous with duration of their appearance varying according to the relative motion of the object and the observer.

From these observations, it is noted that the key to encode these stimuli in neuronal representation is in tracking the changes of the stimulus \cite{hopfield1996transforming}. Such an idea provides us with a new doctrine of neural encoding: can we just encode a finite number of events, as and when key features critical for perception and cognition occur \cite{hopfield2001moment} \cite{hopfield2004encoding}?

\begin{figure}[htbp]
    \centering
    \includegraphics[width=0.38\textwidth]{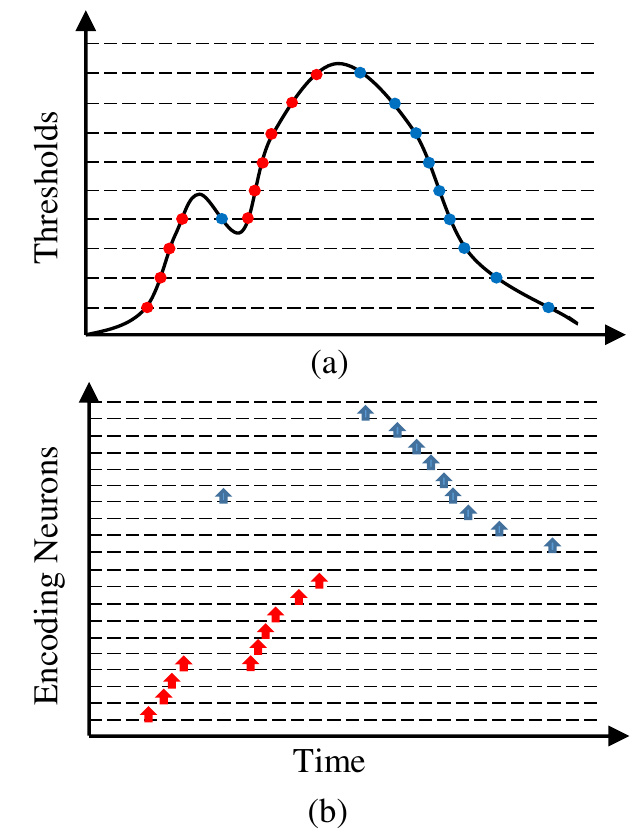}
    \caption{Threshold coding: a snippet of the stimulus (a) and the corresponding twenty encoding neurons (b) that encode the stimulus using ten unique threshold values. The threshold-crossing events are encoded using spikes of neurons with the corresponding thresholds(red refers to onset, blue refers to offset).}
    \label{threshold code}
\end{figure}

Inspired by this idea, the threshold code is proposed as a unique coding scheme that encodes the trajectory of a time-varying signal \cite{gutig2009time}. To encode such a signal, the selectivity of a population of neurons are defined by having their unique firing thresholds set to the corresponding intensity values of the signal. Given a set of intensity values (and hence firing thresholds), the population of neurons is twice as many, as they can be divided into neurons encoding for threshold crossing from beneath (onset neurons) and those for threshold crossing from above (offset neurons), as shown in Figure \ref{threshold code}. In Figure \ref{threshold code} (a), the black curve represents a continuous and slow-varying stimulus, for instance, an acoustic signal from one sub-band cochlear filter. The ten dashed black horizontal lines represent the intensity values which also correspond to the thresholds of the encoding neurons. Red and blue dots refer to onset and offset crossing events respectively. Figure \ref{threshold code} (b) shows the encoded spike patterns (arrows are spike events), with the same color scheme to denote onset and offset neurons, a total of twenty encoding neurons.

\section{Experiments}

\begin{figure}[h!]
    \centering
    \includegraphics[width=0.48\textwidth]{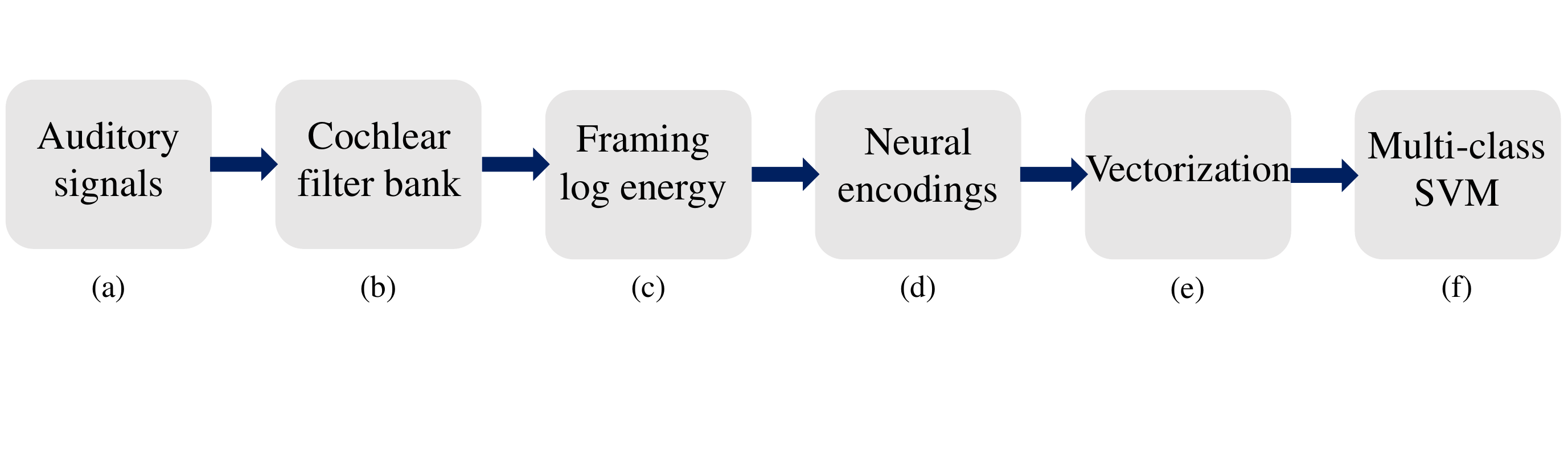}
    \caption{Illustration of steps taken in the experiment. The auditory signals (a) (speech or sounds) are decomposed into multiple channels by the cochlear filter bank in (b) and framed into a spectrogram in (c); the parallel streams of sub-band signal frames are encoded into spike patterns by the different coding schemes in (d). The spatio-temporal spike patterns may then be further vectorized into 1-dimensional vectors in (e), that are then classified by an SVM in (f).}
    \label{block}
\end{figure}

In this section, we investigate how different encoding schemes affect recognition accuracies using a linear SVM classifier. Acoustic signals are good candidates for the experiments, as they are continuous (hence also dynamic), while the spatial dimension refers to the encoding neurons and not object positions in a video, for instance, making it more amenable to recognition using simple classifiers. Compared with static image datasets, they are more similar to real-world stimuli. 

Our study is based on two commonly used acoustic datasets: TIDIGITS \cite{leonard1993tidigits} and RWCP \cite{nakamura2000acoustical}. TIDIGITS is a spoken digit dataset that includes 2,464 training and 2,486 testing utterances, in which each utterance is an isolated spoken digit (`0'-`9',`oh') that spoken by people with different accents. The RWCP is an environmental sound dataset with randomly selected 200 training and 200 testing samples. Each sample contains a sound such as `bell', `phone', `ring', etc. This dataset is relatively small, but offers a different real-world scenario, as natural environmental sounds cover a much wider frequency range than human speech. 

\subsection{Experiment set-up}

\subsubsection{Spectral features and neural encodings}

\begin{figure}[h]
    \centering
    \includegraphics[width=0.45\textwidth]{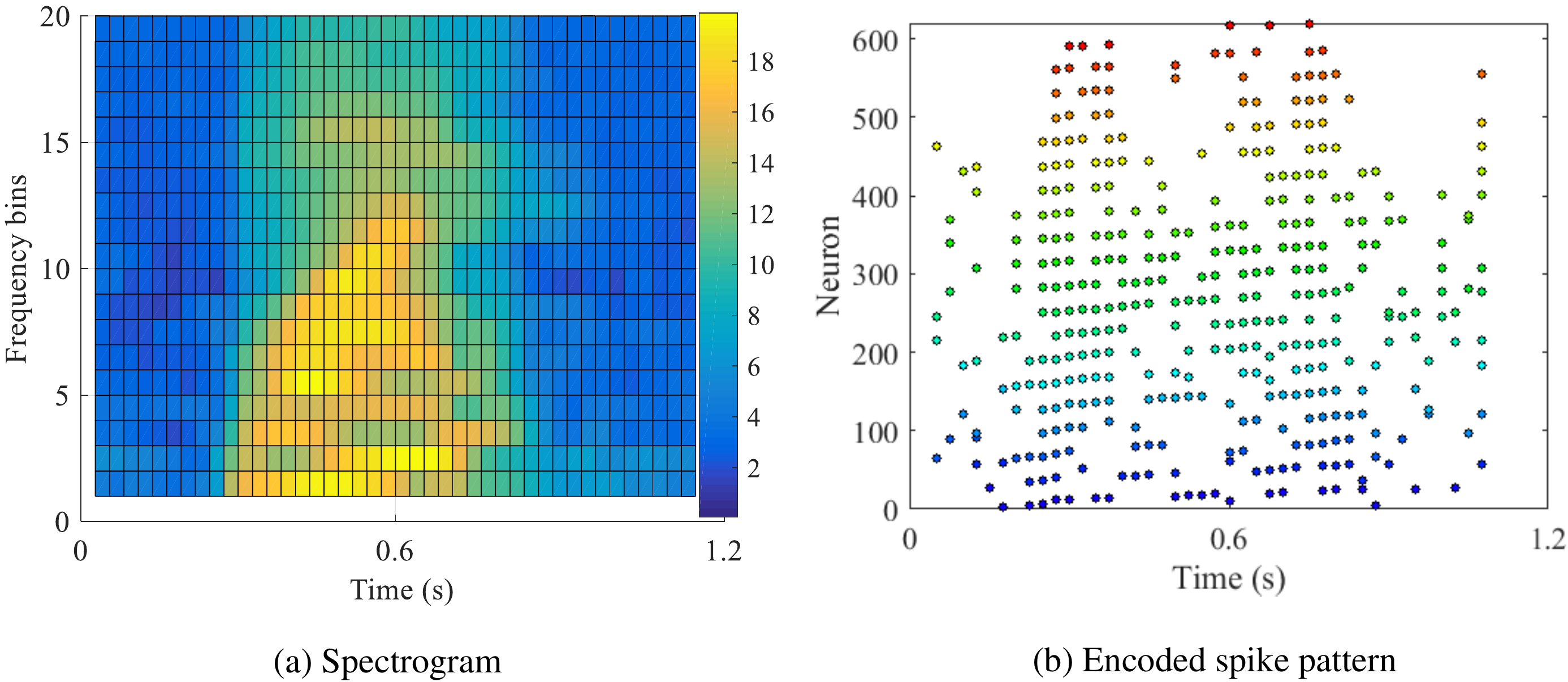}
    \caption{(a) is a spectrogram, a typical output at stage (c) of Figure \ref{block}. Each block represents the spectral energy detected within each sensory window of the hair cells; (b) is an encoded spike pattern using the threshold code (Section   3B), a typical output at stage (d) of Figure \ref{block}. Each dot represents a spike event, with different colors denoting different frequency channels.}
    \label{code_pattern}
\end{figure}

Figure \ref{block} illustrates the steps taken in the experiment. Figure \ref{block} (a), (b) and (c) extract the spectral features of the acoustic signals, as discussed in \cite{pan2018event}. The speech or sound waveform passes through a constant-Q cochlear filter bank with 20 channels. The 20-channel outputs from the filter bank are further framed （with 20ms window size and 10ms stride） and transformed into log energy, emulating the function of auditory hair-cells \cite{muller2008cadherins}. The output of (c) is a two-dimensional spectrogram, as shown in Figure \ref{code_pattern}(a), with 20 rows representing the 20 frequency bins and the number of columns corresponding to the number of time bins used to encode the signal. The spectrograms are further encoded into spike patterns as shown in Figure \ref{code_pattern}(b), using the encoding schemes discussed in Section 3. In this case, we use threshold coding scheme to generate the spike trains.

\subsubsection{Vectorization}
As a linear classifier such as the SVM (Figure \ref{block} (f)) works with only spatial data, the output of Figure \ref{block} (d), a 2-D spatio-temporal spike pattern is further vectorized into a 1-D vector, as in Figure \ref{block} (e). We propose two vectorization techniques:

\begin{figure*}[h]
	\centering
	\includegraphics[width=0.7\textwidth]{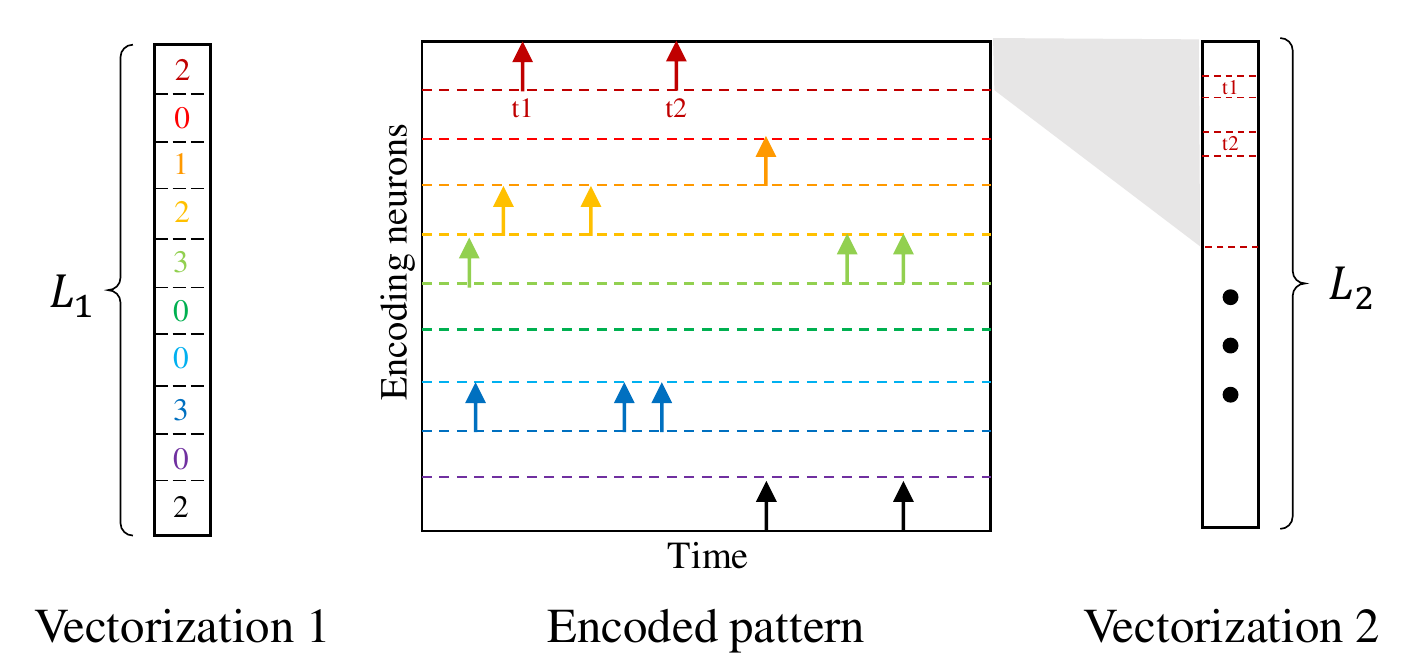}
	\caption{The two-dimensional spatio-temporal spike pattern is converted into a one-dimensional vector using two different approaches. Vectorization 1: the spike train of each encoding neuron is represented by the average firing rate, hence retaining only the spatial information. Vectorization 2: the two-dimensional spike pattern is reshaped into a one-dimensional vector, whereby each value represents the spike timing within each encoding window, hence retaining both spatial and temporal information.}
	\label{vectorize}
\end{figure*}

\begin{enumerate}[1.]
	\item We average the number of spikes for each encoding neuron along the time dimension, using the time-averaged spike rates to represent the output of each encoding neuron, effectively generating an output vector of length $L_1$, equal to the number of neurons. Hence only spatial information is preserved, while temporal information is collapsed. In practice, the duration of each sound pattern is scaled to the mean duration, and the spike counts are used in place of time-averaged spike rates. `Vectorize 1' of Figure \ref{vectorize} illustrates such a vectorization scheme.
	
	\item  We first organize the two-dimension spike patterns into $N_T$ number of time bins by $N_e$ number of encoding neurons. For a given time bin that contains a spike, it is assigned the value of the spike timing; and 0 if there is no spike. This matrix is then row-wise reshaped into a vector, as illustrated by `Vectorize 2' of Figure \ref{vectorize}. The length $L_2$ of `Vectorize 2' is $N_e \times N_T$.  By such a reshaping, both spatial and temporal information is preserved. 
\end{enumerate}

\subsubsection{Classification using SVM}

After vectorization (Figure \ref{vectorize}), the one-dimensional spike patterns are fed into the multi-class SVM. Being a linear classifier, it is useful to assess which neural coding schemes are more effective in projecting spatio-temporal patterns into higher spatial dimensions, and in the process, improve the linear separability of the underlying dataset. The SVM is implemented in the MATLAB ClassificationECOC class, which uses $N$ linear SVM classifiers and adopts a one-against-all strategy for an $N$-class pattern recognition task.

\subsubsection{Classification using SNN}
To further evaluate the effectiveness of these spike encoding schemes, we directly feed the spike patterns into an SNN and trained with the Tempotron learning rule \cite{gutig2006tempotron}. The Tempotron is a temporal learning rule that learns to fire spikes for patterns from the desired class, while remaining silent for patterns from other classes. In our experiments, the number of spiking neurons is equal to the number of sound classes and the one-against-all strategy has been adopted for classification\cite{wu2018spiking}. 

\subsection{Results}

\begin{table*}[htb]
  \begin{center}
    \caption{SVM Classification accuracies (\%) of different neural codings for the TIDIGITS speech dataset}
    \centering
    \label{TIDIGITS SVM}
    \begin{tabular}{c|c|c|c|c|c|c} 
       \hline
     & \multicolumn{3}{c}{Vectorization 1} & \multicolumn{3}{c}{Vectorization 2}\\
    \hline
      \textbf{Neural codings} & \textbf{Vector length} & \textbf{Train Acc of V1} & \textbf{Test Acc of V1}  & \textbf{Vector length} & \textbf{Train Acc of V2}   & \textbf{Test Acc of V2}     \\
      \hline
      Latency coding & 20 & 24.4 & 20.2 & 1640 & 100.0 & 93.7\\
      Phase coding & 20 & 23.1 & 20.6 & 1640 & 100.0 & 92.4\\
      Population latency coding & 200 & 100.0 & 92.7 & 16400 & 100.0 & 92.8\\
      Population phase coding & 200 & 100.0 & 89.8 & 16400 & 100.0 & 91.8 \\
      Threshold coding & 620 & 100.0 & 95.2 & 50840 & 100.0 & 83.4 \\
    \end{tabular}
  \end{center}
\end{table*}

\begin{table*}[htb]
  \begin{center}
    \caption{SVM Classification accuracies (\%) of different neural codings for the RWCP environmental sound dataset}
    \centering
    \label{RWCP SVM}
    \begin{tabular}{c|c|c|c|c|c|c} 
           \hline
     & \multicolumn{3}{c}{Vectorization 1} & \multicolumn{3}{c}{Vectorization 2}\\
    \hline
      \textbf{Neural codings} & \textbf{Vector length} & \textbf{Train Acc of V1} & \textbf{Test Acc of V1}  & \textbf{Vector length} & \textbf{Train Acc of V2}   & \textbf{Test Acc of V2}     \\
      \hline
      Latency coding & 20 & 49.0  & 48.5 & 1480 & 100.0 & 100.0\\
      Phase coding & 20 & 50.5 & 48.0 & 1480 & 100.0 & 100.0\\
      Population latency coding & 200 & 100.0 & 100.0 & 14800 & 100.0 & 100.0\\
      Population phase coding & 200 & 100.0 & 100.0 & 14800 & 100.0 & 100.0\\
      Threshold coding & 620 & 100.0 & 99.5 & 45880 & 100.0 & 99.0 \\
    \end{tabular}
  \end{center}
\end{table*}

\begin{table*}[htb]
 \begin{center}
  \caption{SNN classification accuracies (\%) of different neural codings on the RWCP and TIDIGITS datasets}
  \centering
  \label{SNN_RWCP_TIDIGITS}
  \begin{tabular}{c|c|c|c|c|c|c} 
   \hline
   & \multicolumn{3}{c}{TIDIGITS} & \multicolumn{3}{c}{RWCP}\\
   \hline
   \textbf{Neural codings} & \textbf{Vector length} & \textbf{Train Acc} & \textbf{Test Acc}  & \textbf{Vector length} & \textbf{Train Acc}   & \textbf{Test Acc}     \\
   \hline
   Latency coding & 20 & 9.09 & 9.09 & 20 & 10.0 & 10.0\\
   Population latency coding & 200 & 54.7 & 49.9 & 200 & 100.0 & 99.0\\
   Threshold coding & 620 & 98.5 & 94.9 & 620 & 100 & 99.5 \\
  \end{tabular}
 \end{center}
\end{table*}

The training and testing accuracies using the SVM classifier are obtained for the TIDIGITS and RWCP datasets, as shown in Table \ref{TIDIGITS SVM} and Table \ref{RWCP SVM}, respectively. From these results, we observe that:
\begin{enumerate}[1.]
\item For the single neuron temporal coding schemes, when temporal information is collapsed with vectorization approach 1, the latency and phase codes have the lowest classification accuracies. While with the vectorization approach 2, whereby the temporal information is preserved, classification accuracies increase to a level above 90\% for both datasets. This result implies that both spatial and temporal information is necessary for such temporally rich patterns.\\

\item For the population temporal neuron coding schemes, both population latency and phase codes achieve very high accuracies with both vectorization approaches. This result suggests that with the high redundancy afforded by the population codes, for some datasets, the temporal information is not as critical or even not necessary for good classification accuracies. Also, even if the time information is available, it may not be necessary to preserve their precise timing.\\

\item For the threshold code, which has a sparse representation of the encoded stimulus, it achieves the highest testing accuracy for vectorization approach 1. However, the testing accuracy for vectorization approach 2 is lower than the other coding schemes, which may be due to their large number of encoding neurons (the largest among all, at $N_e=620$), hence generating the longest vectors among all. The 100\% training accuracy suggest good linear separability in the training set, but the SVM has been overfitted during training, given the much lower testing accuracy.\\

\item For the population coding schemes, training accuracies obtained are 100\% for both datasets, using vectorization approach 1. This implies that the training data is linearly separable. While this cannot be generalized to the testing datasets, it nonetheless shows that the population coding schemes have been successful in encoding the spatio-temporal datasets into a spatially linearly separable form.\\
\end{enumerate}


The classification accuracies for the SNN Tempotron learning rule are shown in table \ref{SNN_RWCP_TIDIGITS}. The single neuron temporal code (i.e., latency code) achieves an accuracy of 9.09\% and 10.0\% respectively for the TIDIGITS and RWCP datasets, which are chance probabilities. It is not surprising that the SNN cannot capture the dynamics of the latency encoded patterns as the latency coding scheme encodes each stimulus value in an encoding time window into a spike, generating as many spikes as there are time windows for all input patterns. The numbers of spikes generated are hence the same for all inputs. The Tempotron is then required to learn to classify based on the minute time differences within each time window (due to the latency code). The presence of these time windows is also not known to the SNN, which makes this a very challenging task. For the population latency coded TIDIGITS, the training and testing accuracies increase dramatically, though they are still some way short of those obtained using the SVM. The population latency code improves accuracy by amplifying these minute time differences in the single neuron latency code in the spatial domain by recruiting different populations of encoding neurons with different receptive fields. The threshold coding scheme gives the best accuracies for TIDIGITS, which is only slightly below those obtained using the SVM. The threshold code effectively recruits different sequences of encoding neurons for different time-varying signals which also greatly simplifies the task for the Tempotron. SNN classification on RWCP encoded using the population coding schemes is as good as those done using the SVM. Improved linear separability of population coded spatio-temporal datasets is hence shown to improve classification results obtained using a Tempotron. 

\section{Discussion}

While the RWCP is a smaller dataset than the TIDIGITS, results from experiments performed on it are in agreement with those of TIDIGITS. Hence, going back to the original question: what is an effective neural coding scheme for temporal pattern classification tasks such that the task is made easier for the downstream classifier? To answer this question, we refer to the two categories of temporal neural codes: single neuron code and population code. 

For the single neuron code (Section 2), using vectorization approach 1, the information available with limited spatial dimension (from 20 encoding neurons) is not sufficiently discriminative for a linear classifier.  While the 100\% accuracy, obtained using vectorization approach 2, merely goes to show that if we are able to preserve all spatio-temporal information with extended spatial dimension, the dataset is still linearly separable. Encoding the pixel values of the spectrogram into a spike using the single neuron code, without indicating the boundaries of the encoding time window, is challenging for a spike neural network to classify. This furthers highlights the relevance of the population code.

For the population code (Section 3), both vectorization approaches achieve high training and testing accuracies. This essentially shows that the population codes have encoded sufficient spatial-temporal information in the extended spatial dimension, and with improved sparsity of spikes been generated as shown in Table \ref{spike rate}. The average spike rates per individual neuron in the population codes are much lower than those of the single neuron codes. Notably, the threshold code has the lowest spike rates, for all neurons and also on average (per individual neuron). With the highest coding efficiency, in terms of the average spike rate, the threshold code is ideal for pattern recognition. However, comparing with the other population codes (population latency and phase code), it has lower fidelity, hence, it may not be a good candidate for spike decoding or reconstruction of the original input signal. The low average spike rates generate a sparse pattern which is easier for pattern recognition and also improves energy efficiency in a neuromorphic hardware implementation, where a significant amount of energies is consumed for spike generation and transmission.

\begin{table}[htb]
  \begin{center}
    \caption{Average spike rates (spikes/s) of different neural codings}
    \centering
    \label{spike rate}
    \begin{tabular}{c|c|c} 
    \hline
    Neural codings & for all neurons & per individual neuron\\

      \hline
      Latency coding & 19,000 & 950\\
      Phase coding & 19,000 & 950\\
      Population latency coding  & 56,810 & 284\\
      Population phase coding & 57.940 & 290\\
      Threshold coding   & 7,818 & 12\\
    \end{tabular}
  \end{center}
\end{table}

 From the application point of view, our findings and discussion above could offer a direction on how to choose the neural coding strategies when facing a temporal dynamic pattern recognition task, particularly the speech recognition task. We find that the temporal learning based SNNs are more capable of extracting sparse spatial information from multiple afferents than the dense temporal information from a single afferent. Since the single neuron codes (latency coding, phase coding) cannot preserve sufficient information in spatial dimensions, thus they are not suitable for SNN-based pattern recognition tasks. On the other hand, the population codes (population latency/phase codings, threshold codings) are capable of project the temporal information into spatial dimension (multiple spike trains), they are better candidates for such tasks. Moreover, if we want to design novel strategies of neural codings, the population codings should be a more promising direction.

Our findings are based on utterances of the isolated spoken digit. While it demonstrates certain qualities of the population code (in particular the threshold code) that makes it easier for pattern recognition, it would require a further extension of our experiments to understand whether the results can be generalized to the case of continuous speech, of which the boundaries between words are not clear.

%

\section{Conclusion}
\label{sec:majhead}


We present and study two types of neural coding schemes: the single neuron codes (latency code and phase code) and the population codes (population latency/phase code, and threshold code). Their biological plausibility and possible applications are reviewed and discussed. To evaluate their coding efficacy, a linear classifier, the multi-class SVM is applied to two datasets, namely TIDIGITS and RWCP, encoded using the above schemes. We propose two vectorization approaches to vectorize 2-D spatial-temporal spike patterns. The first approach obtains the time-averaged spike rates for the encoding neurons; while the second maintains all spatial-temporal information. From the experiments, single neuron codes do not contain sufficient information in the spatial dimension for the recognition tasks; while the population codes contain sufficient information in the spatial dimension so as to obtain good classification accuracies, for both the SVM and an SNN-based classifier. The projection of spatial-temporal information onto the spatial dimension using a population code is an intriguing finding that deserves further study. Such a mechanism has a biological basis and showing that such an encoding scheme makes a dataset linearly separable is only the beginning.

\section{Acknowledgments}
This research is supported by Programmatic grant no.
A1687b0033 from the Singapore Government’s Research, In-
novation and Enterprise 2020 plan (Advanced Manufacturing
and Engineering domain)

\bibliographystyle{unsrt}
\bibliography{myreference}



%




\end{document}